# Four Side Distance: A New Fourier Shape Signature

Sonya Eini
Department of Computer Engineering
Razi University
Kermanshah, Iran
sonya.eini@gmail.com

Abdolah Chalechale
Department of Computer Engineering
Razi University
Kermanshah, Iran
chalechale@razi.ac.ir

*Abstract*— **Shape is one of the main features in content based image retrieval (CBIR). This paper proposes a new shape signature. In this technique, features of each shape are extracted based on four sides of the rectangle that covers the shape. The proposed technique is Fourier based and it is invariant to translation, scaling and rotation. The retrieval performance between some commonly used Fourier based signatures and the proposed four sides distance (FSD) signature has been tested using MPEG-7 database. Experimental results are shown that the FSD signature has better performance compared with those signatures.**

*Keywords-Content based Image Retrieval (CBIR); Shape; Four Sides Distance (FSD); Fourier Signatures.*

## I. INTRODUCTION

Because of the increasing amount of digital images and necessity of efficient image retrieval systems, content based image retrieval (CBIR) systems was introduced in the early 1990's [1]. CBIR systems do not have the problems of traditional text based image retrieval (TBIR) systems. In CBIR systems, users' requests are in form of a query image and images are retrieved based on visual contents (features) of images [2]. Color, texture and shape are the main low-level features for CBIR. A block diagram of CBIR systems is shown in Fig. 1. In CBIR systems, a feature vector of each image is extracted based on one or a mixture of those low-level features. Then, the similarity distance between the feature vector of the query image and the feature vectors of the database's images are measured. Finally, system retrieves similar images to the query based on their similarity values.

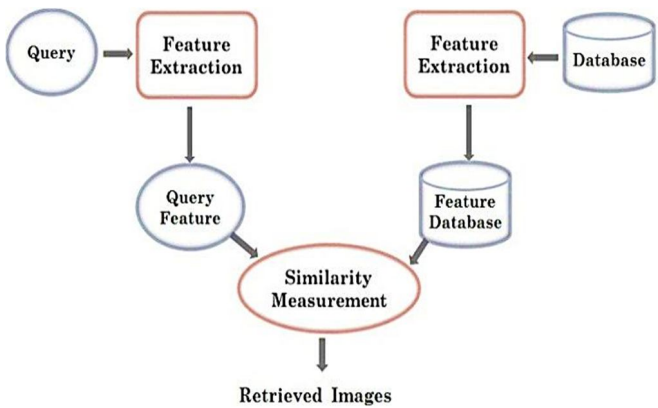

Figure 1. Block diagram of CBIR systems.

In this paper, a new content based image retrieval technique using shape feature is proposed. The proposed shape signature uses the distances between a shape and four sides of the rectangle that covers it.

The rest of the paper is organized as follows: Section II is related works. The proposed signature is presented in Section III. The experimental results are studied in Section IV and Section V is conclusion.

## II. RELATED WORKS

Shape is the most important feature for recognition of objects in an image [3]. There are two classes of techniques in shape based retrieval systems. Region based techniques and boundary based techniques. A region based technique uses whole shape region but a boundary based technique only uses boundary points of shapes in feature vector extraction.

Region based techniques often involve intensive computations and fail to distinguish between objects that are similar [4]. Thus boundary based techniques are more efficient than region based techniques. Several number of techniques have presented that are based on boundary of shapes.

One important class of boundary based techniques is Fourier descriptors (FD). In the FD methods, the Fourier transformed boundary is used as a shape feature [5]. The discrete Fourier transform of a signature $r(t)$ is computed as (1).

$$a_n = \frac{1}{N}\sum_{t=0}^{N-1} r(t) e^{-\frac{j2\pi nt}{N}} \quad n = 0, 1, \dots, N-1 \quad (1)$$

The $a_n$ coefficients are called the Fourier descriptors of the shape. They should be used as (2) to be invariant against translation, scaling and rotation of images.

$$fd = \left[\frac{|a_1|}{|a_0|}, \frac{|a_2|}{|a_0|}, \dots, \frac{|a_{N/2}|}{|a_0|}\right] \quad (2)$$

Some commonly used Fourier based signatures that are used in our experiments, are polar coordinate (PC), complex coordinate (CC), angular radial coordinate (ARC), angular function (AF), triangle area representation (TAR) and chord length distance (CLD).

## A. Polar Coordinate (PC)

The feature vector of this signature at each boundary point $(x_t, y_t)$ is a complex number. The real part is the distance between the point and the centroid $(x_c, y_c)$ of the shape. It is computed as (3). The complex part is the angle between *radial* and *x* axis [8]. See Fig. 2(a). Equation (4) is the feature vector of this descriptor.

$$radial(t) = \sqrt{(x_t - x_c)^2 + (y_t - y_c)^2} \quad (3)$$

$$PC(t) = radial(t) + j\theta(t) \quad (4)$$

## B. Complex Coordinate (CC)

The feature vector of this signature at each boundary point $(x_t, y_t)$ is a complex number. The real part is $(x_t-x_c)$ and the complex part is $(y_t-y_c)$ [7]. See Fig. 2(b). Equation (5) is the feature vector of this descriptor.

$$CC(t) = (x_t - x_c) + j(y_t - y_c) \quad (5)$$

## C. Angular Function (AF)

This signature considers the changes of directions for some boundary points of a shape (with *s* step) [4]. The feature vector of this descriptor is computed as (6). See Fig. 2(c).

$$\varphi(t) = \frac{(y(t) - y(t-s))}{(x(t) - x(t-s))} \quad (6)$$

## D. Angular RadialCoordinate (ARC)

The feature vector of this signature at each boundary point is a complex number that the real part is *radial* and the complex part is same as the AF signature [8]. See Fig. 2(d). Equation (7) is the feature vector of this descriptor.

$$ARC(t) = radial(t) + j\varphi(t) \quad (7)$$

## E. Triangle Area Representation (TAR)

The feature vector of this signature is the area formed by each sequential three boundary points of a shape. It distinguishes between concave and convex regions [9]. See Fig. 2(e).

## F. Chord Length Distance (CLD)

In this signature, the feature vector at each boundary point *p* is the distance between that point and another boundary point *q*, such that *pq* be a vertical line to the tangent vector at *p* [10]. See Fig. 2(f).

## III. THE PROPOSED SIGNATURE

In this section, a new shape signature is proposed. It is a boundary based approach.

In the first stage, the rectangle that covers a shape is obtained. Then, *N* points on the shape's boundary are considered. These selected points should have a same distance with each other. In the most important phase of the feature extraction, four values for each one of the selected boundary points are obtained. These four values for each point are defined as:

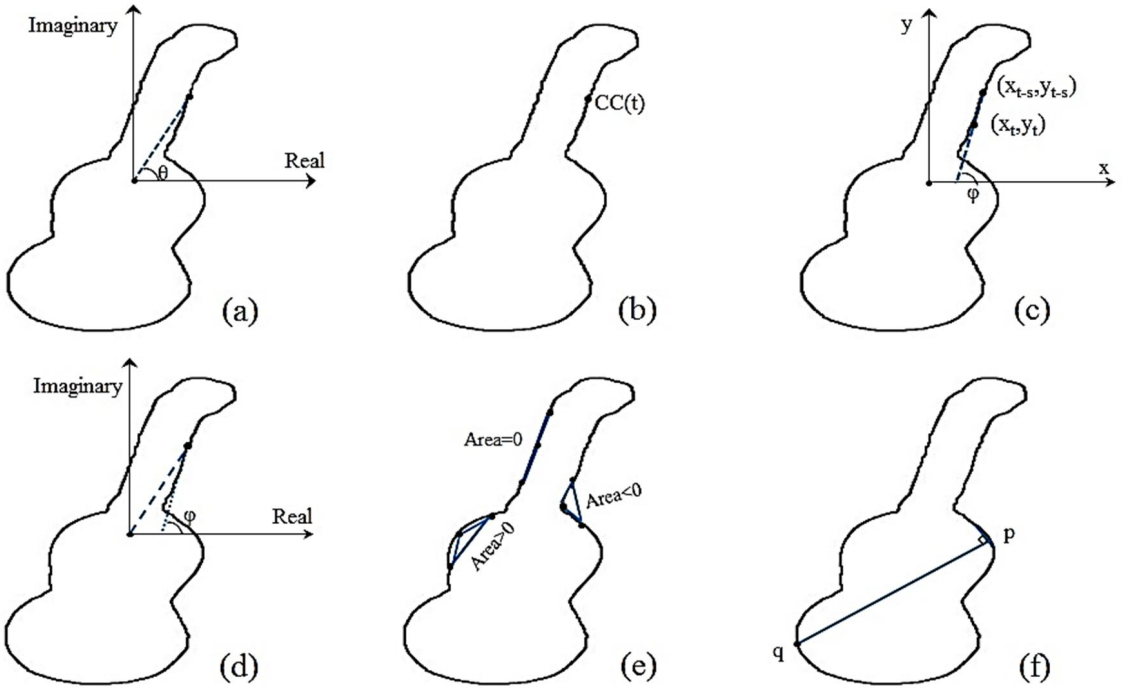

Figure 2. Fourier shape signatures: (a) Polar Coordinate (PC) signature; (b) Complex Coordinate (CC) signature; (c) Angular Function (AF) signature; (d) Angular Radial Coordinate (ARC) signature; (e) Triangle Area Representation (TAR) signature; (f) Chord Length Distance (CLD).

- The distance between the boundary point and the opposite point of it at the top side of the rectangle. This is the first element of this point in feature vector and it is called *dTop*.

- The distance between the boundary point and the opposite point of it at the right side of the rectangle. This is the second element of this point in feature vector and it is called *dRight*.

- The distance between the boundary point and the opposite point of it at the bottom side of the rectangle. This is the third element of this point in feature vector and it is called *dBottom*.

- The distance between the boundary point and the opposite point of it at the left side of the rectangle. This is the fourth element of this point in feature vector and it is called *dLeft*.

Therefore, the size of the proposed method is 4*N*. This signature is called four side distance (FSD). The feature vector for the $t_{th}$ point of the FSD is computed as (8).

$$fv(t) = \{dTop_t, dRight_t, dBottom_t, dLeft_t\} \quad (8)$$

The feature extraction process is displayed in Fig. 3. The original shape is shown in Fig. 3(a). Fig. 3(b) shows the shape and the rectangle that covers it. It also shows that how to compute four values for one of the boundary points of the shape.

Fig. 4 shows FSD signature for two different classes of shapes. It is clear that the FSD feature vectors of same classes of shape are similar to each other.

A proper shape retrieval signature should be invariant against translation, scaling and rotation of the shape. The FSD signature is translation invariant. It uses the smallest rectangle that covers the shape. Extraction of this rectangle is independent from the translation of the shape.

*A. Scale Normalization*

For normalization of the FSD signature against scale, the obtained distances should be normalized. If a distance is from the length side of the rectangle, we should divide it to the width of the rectangle (*W*). If a distance is from the width side of the rectangle, we should divide it to the length of the rectangle (*L*). Therefore, the feature vector of the FSD for $t_{th}$ point changes as (9).

$$fv(t) = \left\{\frac{dTop_t}{W}, \frac{dRight_t}{L}, \frac{dBottom_t}{W}, \frac{dLeft_t}{L}\right\} \quad (9)$$

*B. Rotation Normalization*

The FSD signature is rotation invariant by using the magnitude values of the descriptor and ignoring the phase information. Therefore, the Fourier coefficients of the FSD signature are calculated as (10).

$$a_n = \frac{1}{4N} \sum_{t=0}^{4N-1} FSD(t) e^{\frac{-j2\pi nt}{4N}}$$

$$n = 0, 1, \ldots, 4N-1$$

$$Feature\ Vector = [|a_1|, |a_2|, \ldots, |a_{2N}|] \quad (10)$$

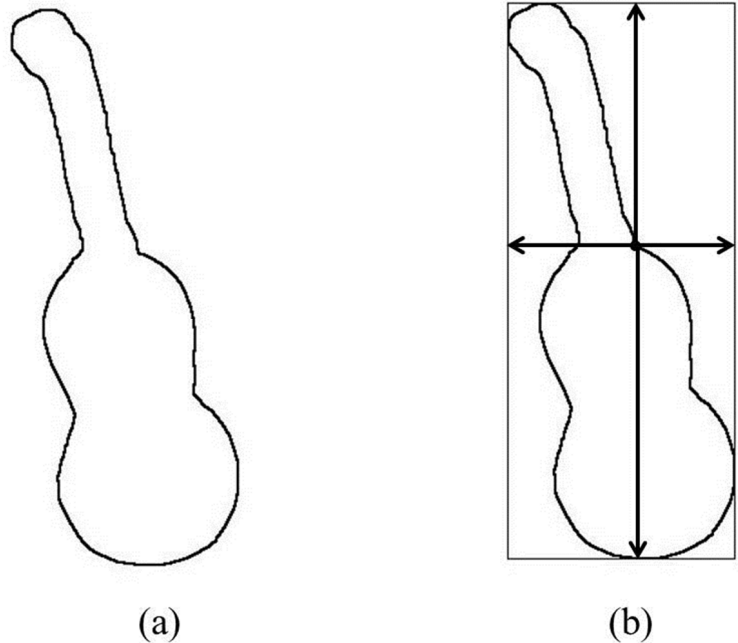

Figure 3. The Four Sides Distance (FSD) signature: (a) The original shape; (b) Computing the distances between the shape's boundary points and four sides of the rectangle that covers it.

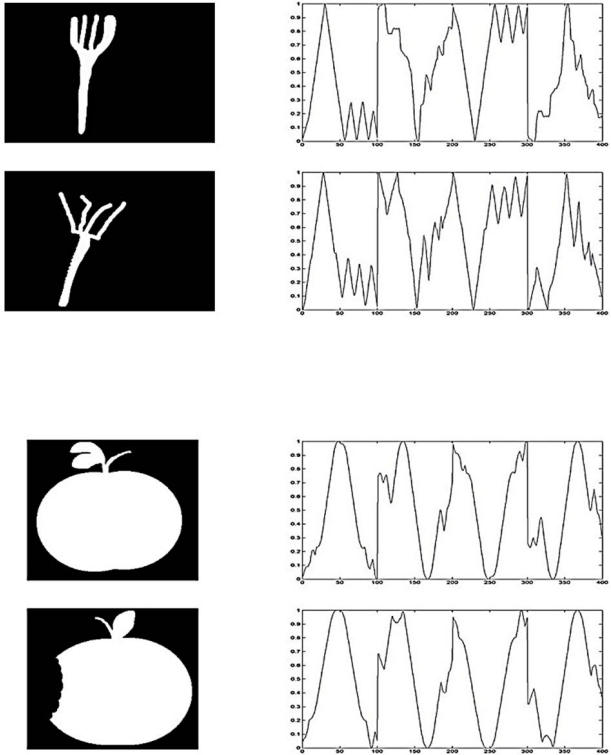

Figure 4. The FSD signatures of two different classes of shapes.

## IV. EXPERIMENTAL RESULTS

For performance measurement and comparing the proposed technique with other commonly used shape signatures, we use part B of the MPEG7 database [11]. It consists of 1400 images that are classified into 70 classes. Each class has 20 similar images. Fig. 5 shows some samples of this database. All the 1400 shapes in the database are used as queries in our experiments.

Euclidean distance has been used for similarity measurement. The retrieval performance is measured in terms of precision and recall. Precision measures the retrieval accuracy, whereas recall measures the capability to retrieve relevant items from the database [12]. They are calculated as (11) and (12), respectively.

$$Precision = \frac{\#\ relevant\ retrieved\ images}{total\ \#\ retrieved\ images} \quad (11)$$

$$Recall = \frac{\#\ relevant\ retrieved\ images}{total\ \#\ relevant\ images\ in\ DB} \quad (12)$$

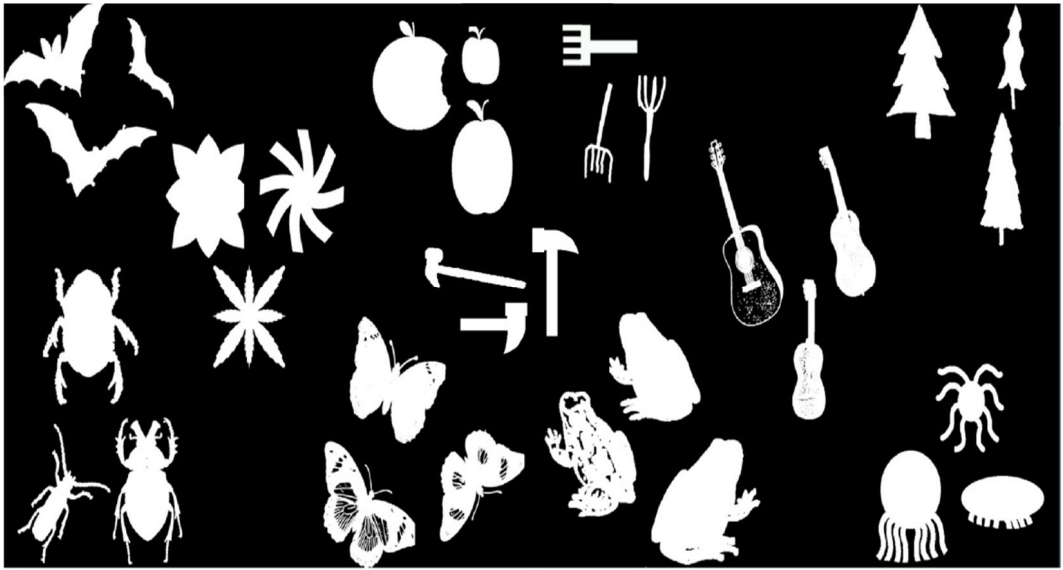

Figure 5. Samples of 11 classes of shapes from set B of the MPEG-7 database.

Some samples of retrieved images based on the FSD signature are shown in Fig. 6. These shapes are in MPEG-7 database. Based on these retrieval shapes, it is clears that the FSD signature is not sensitive against translation, scale and rotation.

Table 1 shows the average of precision for low and high levels of recalls (recall ≤ 50% and recall > 50%) for the FSD signature and seven well-known signatures. It is clear that the performance of the FSD signature is higher than them. For better comparison, Fig. 7 shows the precision-recall curves of the proposed FSD method and two of those shape signatures (CC and CLD). It is obvious that the precision-recall curve of the FSD method is higher than the CLD and CC signatures. It is concluded that, the proposed method has a good performance in comparison with some of well-known shape signatures.

TABLE I. THE AVERAGE PRECISION FOR HIGH AND LOW LEVELS OF RECALL FOR THE FSD AND OTHER FOURIER DESCRIPTORS ON THE MPEG-7 DATABASE.

| Signatures | Average Precision | |
|---|---|---|
| | *Recall ≤ 50%* | *Recall > 50%* |
| FSD | 67.72 | 35.83 |
| PC | 64.40 | 35.12 |
| CC | 64.76 | 22.59 |
| ARC | 58.93 | 26.83 |
| AF | 57.39 | 27.88 |
| TAR | 58.70 | 23.54 |
| CLD | 57.80 | 24.00 |

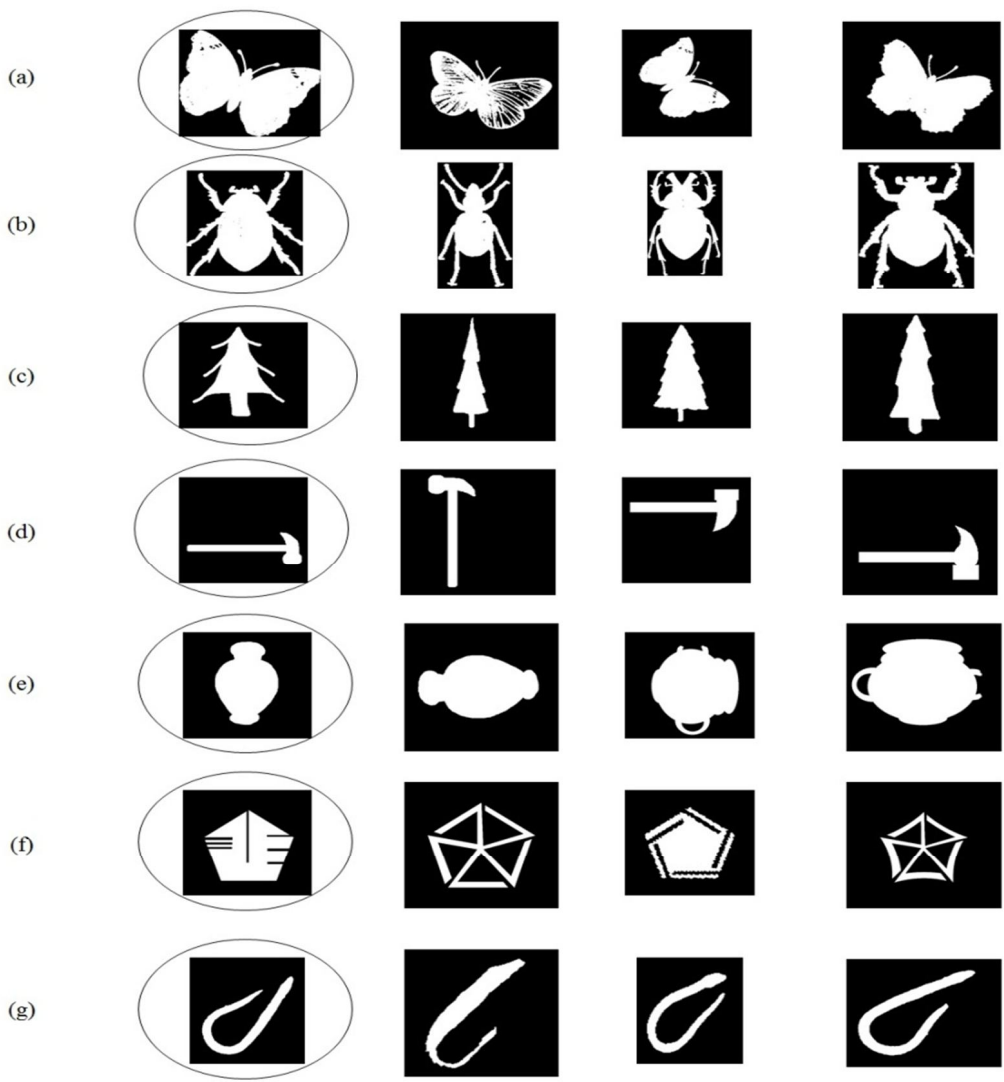

Figure 6. Retrieval images for seven classes of shape in MPEG-7 database based on the FSD signature: (a) Butterfly class; (b) Beetle class; (c) Tree class; (d) Hammer class; (e) Jar class; (f) Device6 class; (g) Sea_snake class.

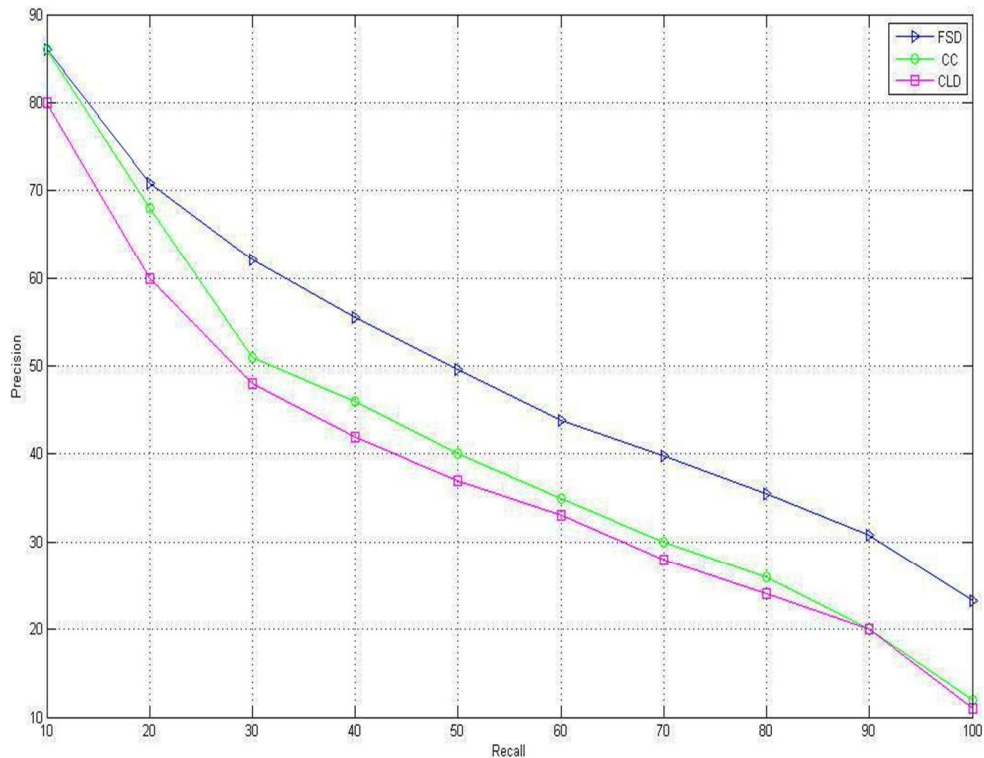

Figure 7. The precision-recall curves of the proposed method (FSD) and some well-known shape signatures.

## V. CONCLUSION

One of the important low-level features in content-based image retrieval is shape. This paper proposed a new CBIR system based on shape feature. This is based on boundary points of shapes. The four sides of the rectangle that covers a shape are considered. Then some points on the surface of the shape, is selected. The distances between the points on the shape's boundary and four sides of the rectangle, are computed as feature vector.

The proposed method that is called four side distance (FSD) is a Fourier based descriptor and it is not sensitive to translation, scaling and rotation. The proposed FSD signature is compared with six commonly used Fourier based descriptors (CC, PC, AF, ARC, TAR and CLD). Experimental results are shown that the average of precision for low and high levels of recall for the FSD signature is higher than those commonly used signatures.

REFERENCES

[1] Suchismita Das, Shruti Garg and G. Sahoo,"Comparison of Content Based Image Retrieval Systems Using Wavelet and Curvelet Transform," The International Journal of Multimedia & Its Applications (IJMA) Vol.4, No.4, pp.137-154, August 2012.

[2] NidhiSinghai and Prof. Shishir K. Shandilya, "A Survey On: Content Based Image Retrieval Systems," International Journal of Computer Applications (0975 – 8887) Volume 4 – No.2, pp. 22-26, July 2010.

[3] N. Alajlan, I. El Rube, M.S. Kamel and G. Freeman, "Shape retrieval using triangle- area representation and dynamic space warping," Patern Recognition, vol. 40, pp. 1911-1920, 2007.

[4] Akrem El-ghazal, OtmanBasir and SaeidBelkasim: "Farthest point distance: Anew shape signature for Fourier descriptors," Signal Processing: Image Communication 24, pp. 572-586, 2009.

[5] R.C. Gonzalez, R.E. Woods, "Digital Image Processing,"Addison-Wesley, Reading, MA, 2002.

[6] K.L. Tan, and L. F. Thiang, "Retrieving similar shapes effectively and efficiently,"Multimedia Tools and Applications, Kluwer Academic Publishers, Netherlands, pp.111–134, 2003.

[7] D.S. Zhang, G. Lu, "A comparative study of curvature scale space and Fourier descriptors," Journal of Visual Communication and Image Representation 14,pp.41–60, 2003.

[8] I.Kunttu and L.Lepisto, "Shape-based retrieval of industrial surface defects using angular radius Fourier descriptor,"IET Image Processing, 2007, 1, (2), pp.231–236, 2007.

[9] Yang Mingqiang, Kpalma Kidiyo and Ronsin Joseph, "A survey of shape feature extraction techniques,"Pattern Recognition, Peng-Yeng Yin (Ed.), pp.43-90, 2008.

[10] D.S. Zhang and G. Lu, "Study and evaluation of different Fourier methods for image retrieval," Image and Vision Computing 23, pp.33–49, 2005.

[11] F. Mokhtarian, M. Bober, "Curvature Scale Space Representation:Theory Application and MPEG-7 Standardization," first ed., KluwerAcademic Publishers, Dordrecht, 2003.

[12] RitendraDatta, DhirajJoshi, Jia Li and James Z. Wang, "Image Retrieval: Ideas, Influences, and Trends of the New Age," ACM Computing Surveys, Vol. 40, No. 2, Article 5, Publication date: April 2008.